%% file: example_paper.tex
\pdfoutput=1
\documentclass[11pt]{article}

\usepackage{acl}

\usepackage{times}
\usepackage{latexsym}
\usepackage{multirow}
\usepackage{xspace}
\usepackage{tcolorbox}
\usepackage{xcolor}
\usepackage{enumitem}
\usepackage[ruled, vlined, linesnumbered]{algorithm2e}
\usepackage{siunitx,makecell}
\usepackage{float}
\usepackage{booktabs}
\usepackage{graphicx}
\usepackage{soul}

\SetCommentSty{mycommfont}
\definecolor{lightorange}{RGB}{245, 237, 211} % Adjust these values as needed

\usepackage[T1]{fontenc}
\usepackage[utf8]{inputenc}

\usepackage{microtype}

\usepackage{inconsolata}
\usepackage{amsmath}
\usepackage{amssymb}
\usepackage{mathtools}
\usepackage{amsthm}
\usepackage{bm}

\definecolor{applegreen}{rgb}{0.55, 0.71, 0.0}
\definecolor{purple}{RGB}{189,181,225}
\definecolor{orange1}{RGB}{249,213,128}
\definecolor{green}{RGB}{176,217,146}
\definecolor{blue}{RGB}{153,185,233}
\newcommand\ours{\textsc{StructChem}\xspace}

\title{Structured Chemistry Reasoning with Large Language Models}

\author{Siru Ouyang\textsuperscript{1}, Zhuosheng Zhang\textsuperscript{2}, Bing Yan\textsuperscript{3}, Xuan Liu\textsuperscript{1},  \\ \textbf{Yejin Choi}\textsuperscript{\textbf{4,5}}, \textbf{Jiawei Han}\textsuperscript{\textbf{1}}, \textbf{Lianhui Qin}\textsuperscript{\textbf{5,6}} \\
   \textsuperscript{1}University of Illinois Urbana-Champaign, \textsuperscript{2}Shanghai Jiao Tong University\\
  \textsuperscript{3}New York University,
   \textsuperscript{4} University of Washington\\
     \textsuperscript{5}Allen Institute for Artificial Intelligence, \textsuperscript{6}University of California San-Diego\\
     \normalsize\texttt{siruo2@illinois.edu} \\}
\begin{document}
\maketitle

\begin{abstract}
Large Language Models (LLMs) excel in diverse areas, yet struggle with complex scientific reasoning, especially in the field of chemistry. Different from the simple chemistry tasks (e.g., molecule classification) addressed in previous studies, complex chemistry problems require not only vast knowledge and precise calculation, but also compositional reasoning about rich dynamic interactions of different concepts (e.g., temperature changes).
% an understanding to dynamic environmental factors like temperature changes.
Our study shows that even advanced LLMs, like GPT-4, can fail easily in different ways. Interestingly, the errors often stem not from a lack of domain knowledge within the LLMs, but rather from the absence of an effective reasoning {\it structure} that guides the LLMs to elicit the right knowledge, incorporate the knowledge in step-by-step reasoning, and iteratively refine results for further improved quality. On this basis, we introduce \ours, a simple yet effective prompting strategy that offers the desired guidance and substantially boosts the LLMs' chemical reasoning capability. 
% Moreover, \ours can integrate existing reasoning approaches and tools, such as program-of-thoughts for precise scientific calculation, which further enhances the results. 
Testing across four chemistry areas---quantum chemistry, mechanics, physical chemistry, and kinetics---\ours substantially enhances GPT-4's performance, with up to 30\% peak improvement. Our analysis also underscores the unique difficulties of precise grounded reasoning in science with LLMs, highlighting a need for more research in this area. Code is available at \url{https://github.com/ozyyshr/StructChem}.

\end{abstract}

\input{sections/introduction}
\input{sections/related_work}

\input{sections/method}
\input{sections/experiment}

\input{sections/analysis}

\section{Conclusion and Discussion}

This paper introduces \ours, a new reasoning structure that guides LLMs to solve complex chemistry problems. \ours explicitly decomposes the reasoning into three critical phrases, including {\it formulae generation} by LLMs that offers the basis for grounded reasoning, {\it step-by-step reasoning} that makes derivations with the identified formulae for a preliminary answer, and {\it confidence-based review-and-refinement} that steers LLMs to progressively revise the previous phases, leading to the final high-confidence answer. Extensive experiments on four datasets of complex chemistry problems from different subfields of chemistry show that \ours significantly boosts the chemistry reasoning capability of different LLMs. In addition, we finetune smaller LMs (e.g., Vicuna-13B) using the generated reasoning from our approach with GPT-4 and obtain strong improvement. Future work could continue to investigate incorporating external, up-to-date knowledge sources and performing retrieval to ensure the quality of the formulae generation. Or designing strategies to transfer and distill chemistry reasoning knowledge from LLMs to smaller LMs.

\section*{Impact Statement}

The impact of this paper lies in its significant contribution to improving the ability of Large Language Models (LLMs), like GPT-4, in the domain of complex chemistry reasoning. The introduction of \ours not only elevates the LLMs' performance in chemical reasoning tasks but also demonstrates the potential to integrate with other reasoning tools and strategies, thereby pushing the boundaries of what artificial intelligence can achieve in the realm of scientific inquiry.
This advancement opens up new avenues for using LLMs in scientific research, education, and industry, where they can assist in acting as teaching agents, conducting experiments, or even in developing new materials and drugs. Furthermore, by highlighting the ongoing challenges and the need for further research in precise, grounded reasoning with LLMs, the paper sets a new direction for future work in AI and machine learning, encouraging a deeper investigation into how AI can more effectively mimic human-like reasoning in scientific domains. Overall, we do not foresee any major risks or negative societal impacts of our work. All the datasets we experiment with are publicly available online. We followed the licenses when conducting experiments on publicly available datasets and human annotations. We will open-source this project upon acceptance to facilitate future research, especially for small research groups or institutions with relatively fewer resources of LLMs.

\bibliography{example_paper}
\bibliographystyle{icml2024}

\clearpage
\appendix
\input{appendix}

\end{document}

%% file: sections/introduction.tex
% Three vital points are incorporated in this problem, entailing the identification of two versions for \textit{vibrational partition function} and their relations.
\section{Introduction}\label{sec:intro}

\begin{figure}[t]
%\vspace{-3mm}
    \centering
    \includegraphics[width=0.95\linewidth]{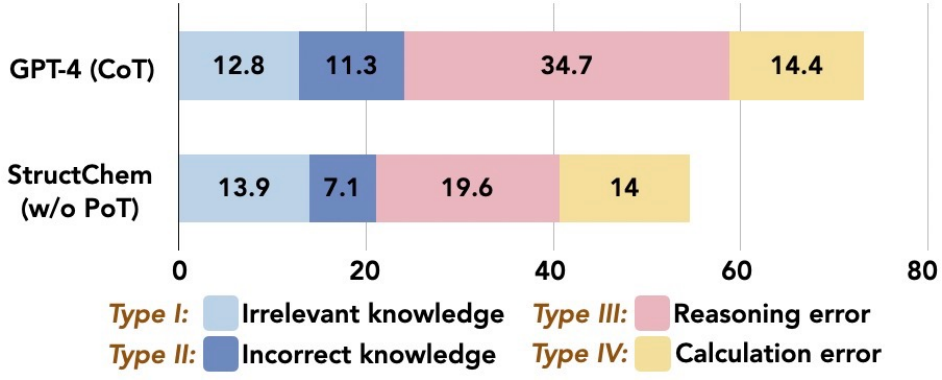}
    % \vspace{-5pt}
    \caption{Proportions (\%) of four error types ( \#errors / \#all-cases) for GPT-4 and \ours. % \ours \textit{w/o PoT} excludes the influence of external tools.
    \ours substantially reduces reasoning error.
    }
    \label{fig:intro-error}
    % \vspace{-5mm}
\end{figure}

\begin{figure*}[t]
%\vspace{-3mm}
    \centering
    \includegraphics[width=0.99\linewidth]{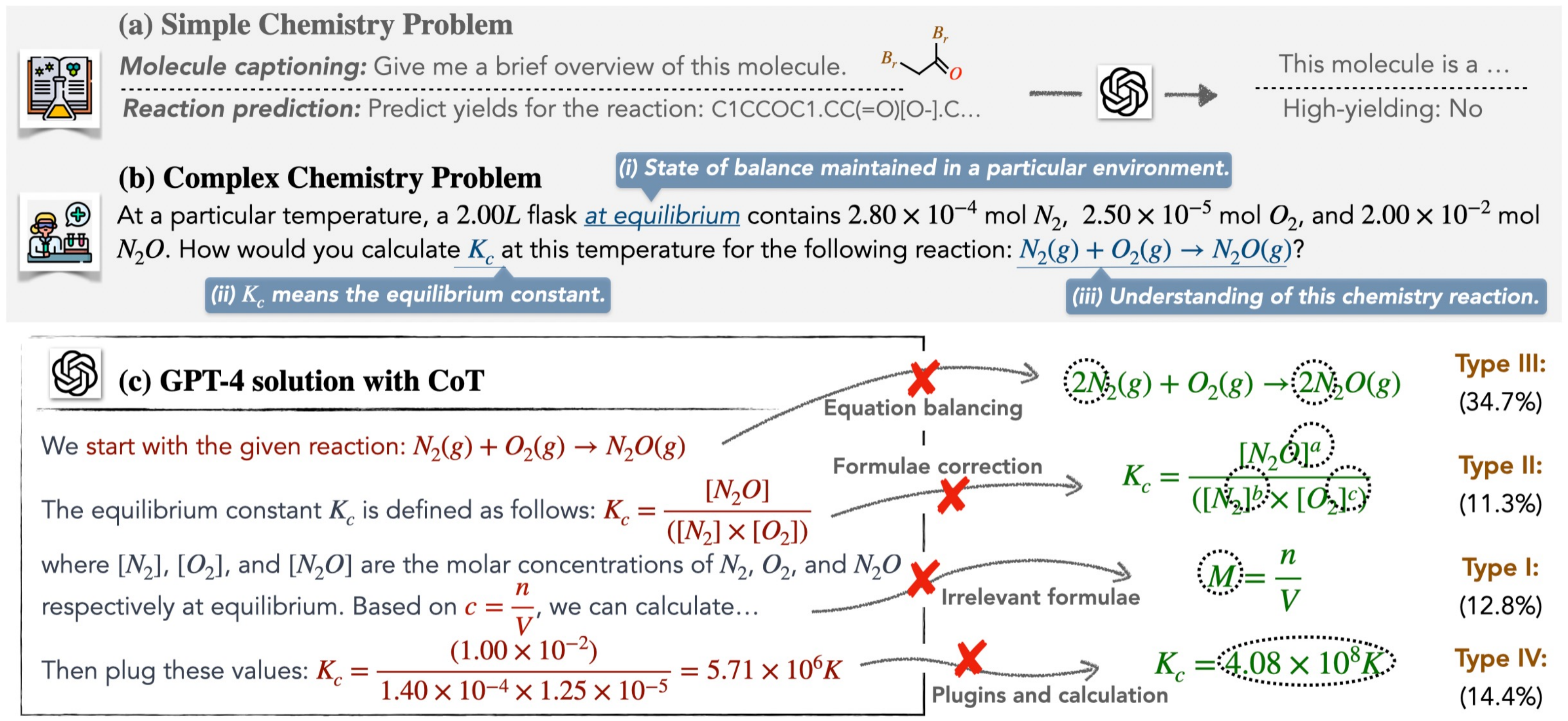}
    \caption{The illustration of \textit{(a) simple chemistry problem}, \textit{(b) complex chemistry problem} sampled from SciBench~\cite{wang2023scibench}, and \textit{(c) the zero-shot response from GPT-4} with chain-of-thought (CoT)~\cite{DBLP:conf/nips/Wei0SBIXCLZ22} for the complex chemistry problem. The error types are illustrated corresponding to the definition in Figure~\ref{fig:intro-error}: (I) irrelevant knowledge, (II) incorrect knowledge, (III) reasoning error, (IV) calculation error. We randomly select 100 error cases of GPT-4 (CoT) in SciBench.}
    \label{fig:intro}
    % \vspace{-3mm}
\end{figure*}

% \noindent Traditional chemistry problems, As illustrated in Figure~\ref{fig:intro}, 

% solving chemistry problems requires more than just language understanding, but also precise domain knowledge (e.g., chemical formulae) and grounded integrative reasoning that connects diverse knowledge and concepts (e.g., temperature, pressure, their interactions) to derive the final answer. Crucially, LLMs must ensure each step of its reasoning is accurate and adheres to the chemistry rules.

Artificial intelligence (AI) holds the promise of transforming the field of chemistry~\cite{baum2021artificial}, impacting various sectors including industrial production~\cite{ozturk2020exploring}, pharmaceuticals~\cite{singhal2023large}, and education~\cite{graulich2022read}. Recent studies have shown promising results of large language models (LLMs) solving simple chemistry problems (Figure~\ref{fig:intro}a), such as molecule classification~\cite{edwards2022translation} and property prediction~\cite{yang2019analyzing, feinberg2018potentialnet}. 

On the other hand, however, more complex chemistry reasoning problems still pose significant challenges to frontier LLMs like GPT-4. As shown in Figure~\ref{fig:intro}b, a complex problem requires not only understanding individual concepts (e.g., molecule property) as in previous tasks, but rather their rich dynamic interactions in different contexts, involving extensive domain knowledge (e.g., chemical formulae\footnote{``Formulae" and ``formulas" are both correct plurals of ``formula'', with ``formulae'' being preferred in scientific writing, per Garner's Modern English Usage.}), precise scientific computing, and compositional step-by-step reasoning. 
As a result, LLMs are prone to different forms of errors when solving these problems, such as applying incorrect knowledge, making miscalculations, or following flawed reasoning processes, as illustrated in Figure~\ref {fig:intro}c.

Interestingly, as shown in Figure~\ref{fig:intro-error} and discussed in Section~\ref{sec:analysis}, LLMs oftentimes have encoded necessary knowledge for a given chemistry problem. The key shortfall, however, lies in the absence of a sophisticated reasoning structure that helps elicit the relevant knowledge from the LLMs, and guides them to perform precise step-by-step reasoning with the knowledge. 

Motivated by this, we introduce \ours, a simple yet effective reasoning strategy providing structured guidance for LLMs to solve complex chemistry problems.  
\ours explicitly decomposes the reasoning into three phases: 
In the {\it first} phase, the LLM focuses on generating essential chemical formulae needed for the problem. The formulae knowledge provides a solid basis for the LLM to do grounded reasoning in subsequent phases. 
The {\it second} phase involves the LLM conducting a detailed, step-by-step reasoning based on the identified formulae, leading to a preliminary answer to the problem.
% {\it Second}, the LLM performs step-by-step reasoning based on the identified formulae and produces a preliminary answer to the problem. 
The {\it third} phase then performs {\it confidence-based} review-and-refinement for the final answer. Crucially, the refinement process differs from recent self-verification methods \cite{DBLP:journals/corr/abs-2303-17651, weng2022large} which rely solely on prompting and can sometimes yield unreliable results (Section~\ref{sec:analysis}). Instead, our approach explicitly estimates a confidence score for each revision, and iteratively enhances the confidence level towards a final high-quality answer.

We conduct extensive experiments on four datasets of complex chemistry problems from different subfields, namely, quantum chemistry, quantum mechanics, physical chemistry, and chemistry kinetics. 
Experiments show that \ours greatly reduces the reasoning errors (Figure~\ref{fig:intro-error}). It boosts the chemistry reasoning capability of advanced LLMs, including GPT-3.5 and GPT-4, leading to an average improvement of 8\% and a 30\% absolute improvement at maximum. 
% \lianhui{mention more analysis}
In addition, using the generated reasoning from our approach with GPT-4, we finetune smaller LMs (Llama-2-13B and Vicuna-13B) and obtain strong improvement. This further validates \ours enables LLMs to generate high-quality chemistry reasoning. Our analysis studies the error patterns of LLMs on chemistry reasoning, which reveals the unique challenges in scientific problems and motivates future research towards more grounded and precise reasoning.

%% file: sections/related_work.tex
\section{Related Work}

% \lianhui{Maybe two subsections: One is ``LLM for Chemistry'' which introduces the existing chemistry problems and datasets and previous AI/LLM methods. This section is for background of chemistry and previous AI/LLM attempts. Highlight that previous LLM for Chemistry are for simple chemistry problems. The second section is LLM reasoning methods (not specific to Chemistry), and highlight our differences.}

\subsection{Large Language Models for Chemistry}

The emergence of LLMs has provided new possibilities in scientific domains, where a bunch of new benchmarks~\cite{lu2022learn, chen2023theoremqa} have emerged. As an important and challenging branch of scientific domains, chemistry-related research surges with the utilization of LLMs~\cite{fang2023mol,tang2024prioritizing,liao2024words}. Specifically, ChemCrow~\cite{bran2023chemcrow} is a general model that integrates multiple existing tools with LLMs to solve various downstream tasks. LLMs are also used to boost the performance of specific chemistry applications, such as reaction prediction~\cite{zhong-etal-2023-reactie}, drug discovery~\cite{edwards2023synergpt}, and SMILES identification~\cite{edwards-etal-2021-text2mol}. However, most previous works target problems that require a single-hop retrieval of domain knowledge, without complex reasoning steps inherent. For example, find and list all the chemical entities in a given sentence. While previous models excel in these approximate knowledge retrieval tasks, improving the abilities of LLMs to solve complex chemistry problems is still in a nascent stage, with SciBench~\cite{wang2023scibench} being the initial benchmark.

% \siru{how to place TheoremQA, chemistryQA, ScienceQA?}

\begin{figure*}[t]
    \centering
    \includegraphics[width=0.99\linewidth]{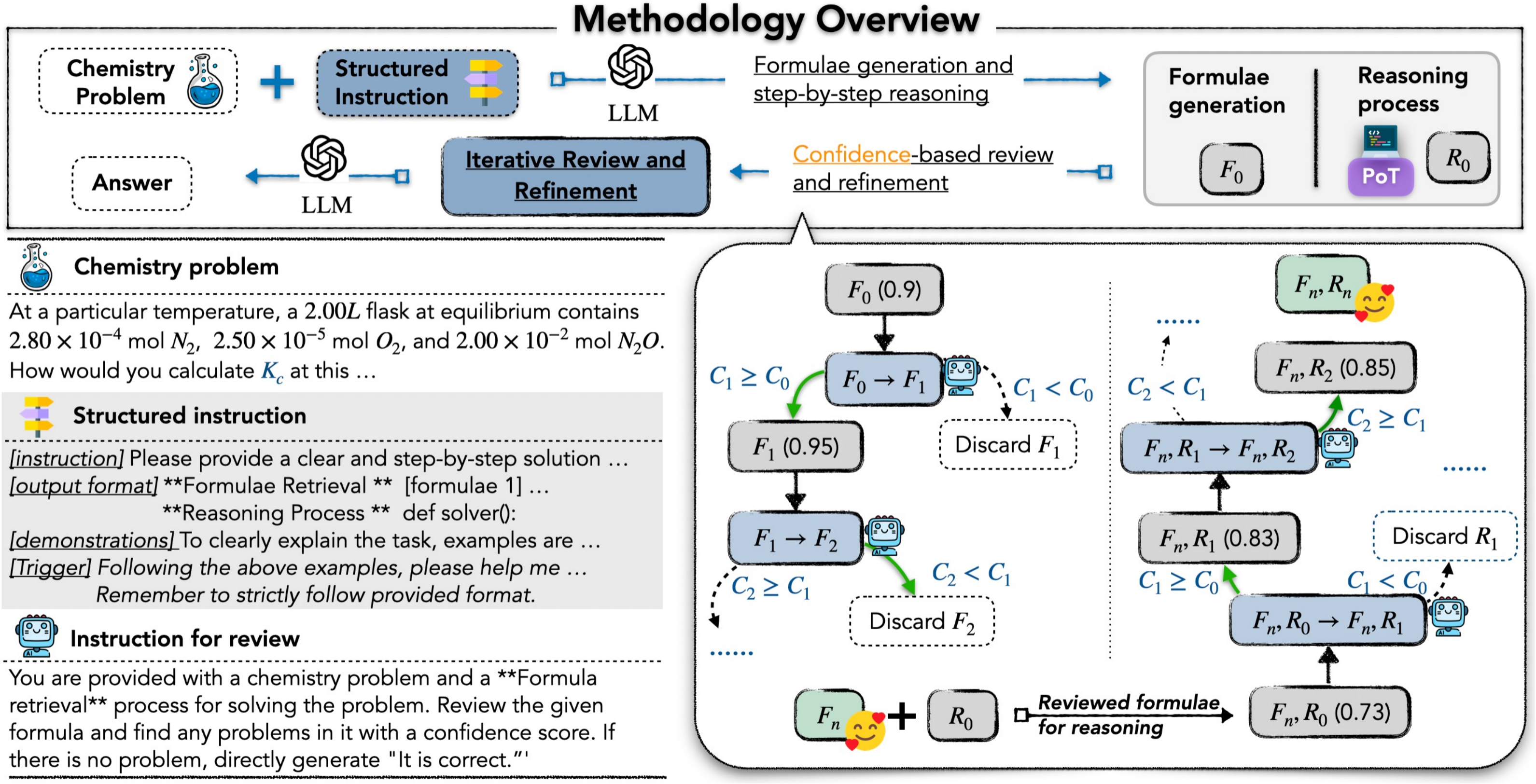}
    \caption{An in-depth illustration of the \ours framework. When tackling a chemistry problem, we first utilize a structured instruction approach, resulting in ``formulae generation'' $\mathcal{F}_0$ and ``step-by-step reasoning'' $\mathcal{R}_0$. These generated segments are then fed to a thorough ``confidence-based review-and-refinement'' as initial input. The process is repeated $n$ times til getting reviewed formulae $\mathcal{F}_n$ and reasoning $\mathcal{R}_n$. Each iteration is guided by incorporating confidence scores $C_i$. \textcolor{applegreen}{\bm{$\rightarrow$}} in ``iterative review and refinement" denote the choice made for each iteration. Full instructions can be found in Figure~\ref{fig:prompt}.}
    \label{fig:method}
    % \vspace{-3mm}
\end{figure*}

\subsection{Large Language Models for Reasoning}

Engaging LLMs in a step-by-step thinking process has demonstrated enhanced performance in intricate reasoning tasks compared to the conventional single-step answer prediction. A typical step-by-step prompting approach is called chain-of-thought (CoT)~\cite{DBLP:conf/nips/Wei0SBIXCLZ22}. CoT directs the model to articulate the step-by-step thinking process as rationales prior to producing the final answer. Following this line, several optimized endeavors are made towards making LLMs better task solvers for more complicated problems. These efforts encompass automated demonstration construction~\cite{zhang2023automatic}, improving self-consistency~\cite{wang2023selfconsistency}, utilizing the structure of prompts~\cite{yao2023tree, besta2023got}, adopting iterative prompting~\cite{zhou2023leasttomost, wang-etal-2022-iteratively}, and ensembling~\cite{sun2023recitationaugmented, fu2023complexitybased}. The original focus of CoT approaches has primarily centered around arithmetic, commonsense, and logical reasoning problems~\cite{DBLP:conf/nips/Wei0SBIXCLZ22,kojima2022large,zhang2023automatic}. Recent investigations have sought to broaden the application of CoT in scientific domains~\cite{lu2022learn,wang2023scibench}. Closely related to ours are works in the research line of modular prompting~\cite{DBLP:conf/iclr/KhotTFF0CS23, patel-etal-2022-question}, which decomposes a complex task into several sub-tasks. The micro-level decomposition varies from task to task. Different from them, we identified two fundamental components for solving complex chemistry problems as a general paradigm at a macro level. 
Another related line is the feedback mechanism~\cite{DBLP:journals/corr/abs-2303-17651} that leverages feedback from LLMs before the final output. In contrast, we design a confidence-based review-and-refinement strategy and employed another LLM to provide feedback for multi-model collaboration. Notably, this approach will greatly alleviate the drawbacks of previous feedback frameworks, where correct answers risk being swayed by unfaithful feedback. 
% \vspace{-1mm}

%% file: sections/method.tex
\section{\ours Reasoning}\label{sec:methdology}

% Our approach, \ours, explicitly decomposes the reasoning into {\bf three phases}: 
% In the {\it first} phase, the LLM focuses on generating essential chemical formulae needed for the problem of interest. The formulae provide a solid basis for the LLM to do grounded reasoning in subsequent phases. 
% The {\it second} phase involves the LLM conducting a detailed, step-by-step reasoning based on the identified formulae, leading to a preliminary answer to the problem.
% % {\it Second}, the LLM performs step-by-step reasoning based on the identified formulae and produces a preliminary answer to the problem. 
% The {\it third} phase then performs iterative review and refinement for the final high-confidence answer.

Solving complex chemistry problems not only necessitates recognizing domain knowledge, such as formulae and calculations but also demands the ability to construct a careful step-by-step reasoning process based on the relevant knowledge.
Existing popular reasoning methods, such as CoT and self-consistency, though exhibit notable strengths, often fall short in accurately identifying the related chemistry formulae and are susceptible to errors in reasoning steps, as expounded in Section~\ref{sec:intro}.

To address the challenges above, we propose \ours.
On a high level, \ours consists of three stages: (i) formulae generation that offers the basis for subsequent grounded reasoning; (ii) step-by-step reasoning that makes multi-step derivations with the identified formulae for a preliminary answer; and (iii) confidence-based review-and-refinement that steers LLMs to progressively revise the previous phases for increasing confidence, leading to the final high-confidence answer. 
Figure~\ref{fig:method} the overall framework. 
% On a high level, \ours consists of three stages: (i) formulae generation; (ii) step-by-step reasoning with formulae; and (iii) iterative review-and-refinement.  The overall framework is shown in Figure~\ref{fig:method}. Firstly, the LLM generates essential chemical formulae needed for the problem. The formulae provide a solid basis for the LLM to perform grounded reasoning in subsequent phases. Secondly, the LLM conducts a detailed, step-by-step reasoning based on the identified formulae, leading to a preliminary answer to the problem. Thirdly, the LLM performs iterative review-and-refinement on the formulae and reasoning steps and derives the final result with a high-confidence score. 

% iterative review and refinement with model collaboration. We first explicitly decompose the overall generation into two high-level parts, formulae collection and reasoning, under the guidance of a pre-defined template. We then employ iterative review and refinement with the help of another model based on the confidence of generated results.
% \vspace{-3mm}
\subsection{Formulae Generation}
\label{sec: phase1}

% Solving complex chemistry problems requires not only identifying domain knowledge such as formulae and calculations but also generating a thoughtful step-by-step reasoning process on top of the concerned knowledge. Although the prevalent CoT prompting strategy appears to be effective with its distinct strengths such as the clarity of explanations, it still has drawbacks for solving such complex chemistry problems~\cite{turpin2023language}. Specifically, the required domain knowledge is often entangled with the reasoning process, making it hard to trace and solve the errors accordingly. Motivated by this, we propose explicitly disentangling the generation process into ``formulae generation'' and ``step-by-step reasoning''.
% The rationale is that grounding the reasoning on top of such domain knowledge abstractions helps to avoid reasoning errors in the intermediate steps, as an example shown in Figure~\ref{fig:intro} (left) from CoT prompting.

\begin{figure}[t]
    \centering
    \includegraphics[width=0.98\linewidth]{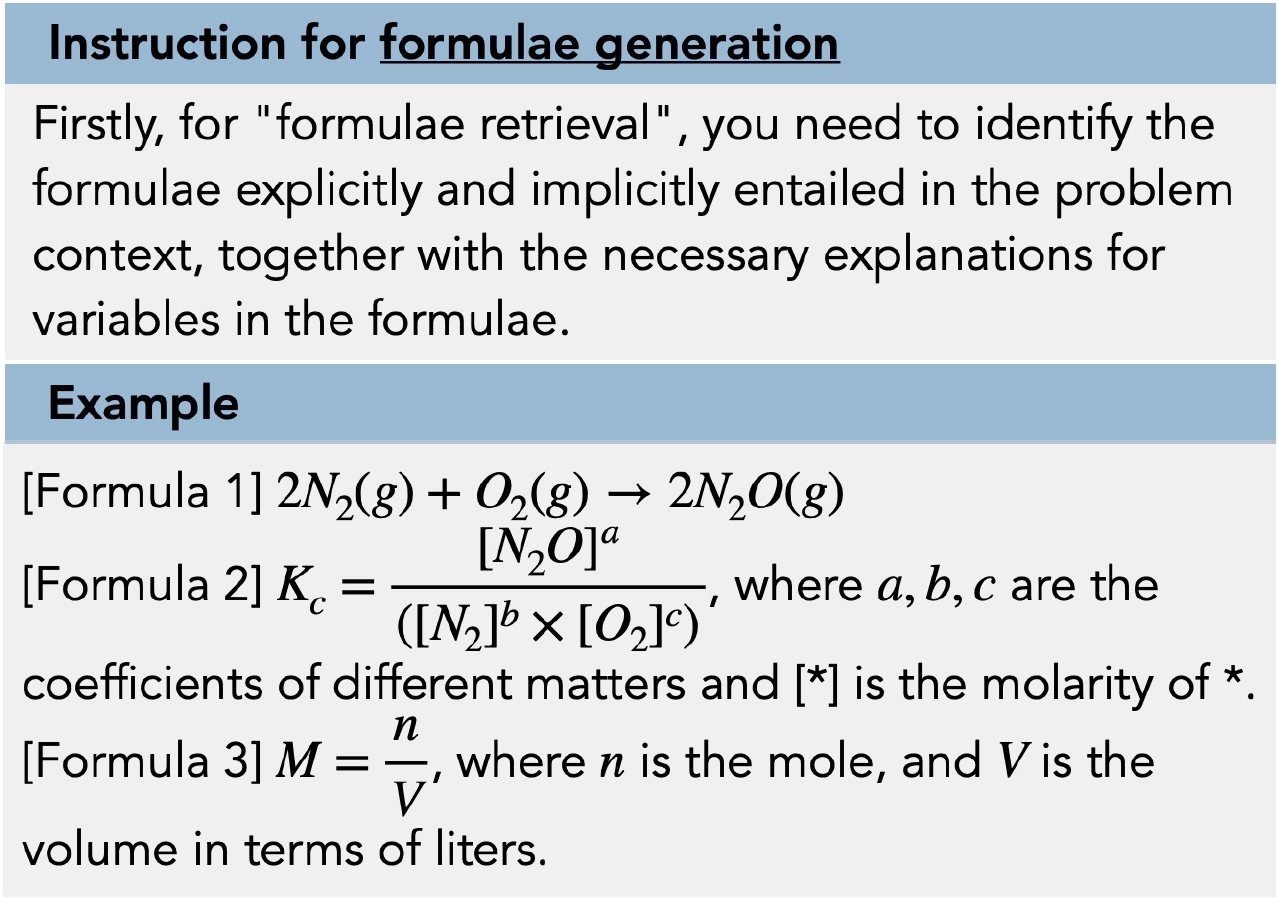}
    \caption{Instruction for formulae generation (part of the \textit{structured instruction} in Figure~\ref{fig:method}) and the output for problem in Figure~\ref{fig:intro}.}
    \label{fig:formulae}
    % \vspace{-6mm}
\end{figure}

Formulae serve as organized and abstracted representations of chemistry knowledge~\cite{lachmy-etal-2022-draw}. When humans tackle intricate problems, the initial phase often involves seeking relevant knowledge as a foundation, especially for the field of chemistry~\cite{taskin2014students}. Therefore, rather than directly starting to address the question, we seek formulae to solve the problem first. Given the fact that LLMs have indeed encoded much chemistry knowledge, it is often effective to elicit the knowledge from the parametric storage~\cite{petroni-etal-2019-language}. Therefore, \ours first instructs the LLM to articulate relevant formulae for task resolution, exemplified by formulae like ``\textit{equilibrium constant}'' in Figure~\ref{fig:formulae}. To enhance the utility of these formulae in subsequent reasoning processes, we instruct the LLM not only to recite them but also to provide explanations for the variables they contain. For instance, as illustrated in Figure~\ref{fig:intro}, the LLM needs to elucidate \textit{symbol $[*]$} as the molar concentrations.

\begin{algorithm}[t]
\small
\caption{Confidence-based Review-and-Refinement}\label{Alg: Iterative Review}
\KwIn{$S_0 = (\mathcal{F}_0, \mathcal{R}_0)$, initial confidence score $C_0^f$ and $C_0^r$, LLM $\mathcal{M}$, prompts $\{p_{rev}, p_{gen}\}$, maximum iteration number $n$}
\KwOut{Final answer to the problem $P$}
\BlankLine
$\mathcal{F}_0 = \{f_1,...,f_n\}$, $\mathcal{R}_0 = \mathcal{R}_0$ \\
\BlankLine
\tcp{initialize maximum confidence scores for iteration}
$max_f\leftarrow C_0^f$, $max_r\leftarrow C_0^r$ \\
\BlankLine
\tcp{review for collected formulae}
\For{$i$ in $1,...,n$}
{
    $(\mathcal{F}_i, C_i^f)\leftarrow \mathcal{M}(p_{rev}||\mathcal{F}_{i-1})$\\
    \If{$C_i^f< max_{f}$}{continue}
    $\mathcal{F} = \mathcal{F}_i$, $max_f\leftarrow C_i^f$
}

\tcp{review for reasoning process}

\For{$i$ in $1,...,n$}
{
    $(\mathcal{R}_i, C_i^r)\leftarrow \mathcal{M}(p_{rev}||\mathcal{F}||\mathcal{R}_{i-1})$\\
    \If{$C_i^r< max_{r}$}{continue}
    $\mathcal{R} = \mathcal{R}_i$, $max_r\leftarrow C_i^r$
}
\tcp{combine reviewed parts}
\Return{$\mathcal{M}(p_{gen}||\mathcal{F}||\mathcal{R})$}
\end{algorithm}

\subsection{Step-by-step Reasoning} 
\begin{figure}[t]
    \centering
    \includegraphics[width=0.98\linewidth]{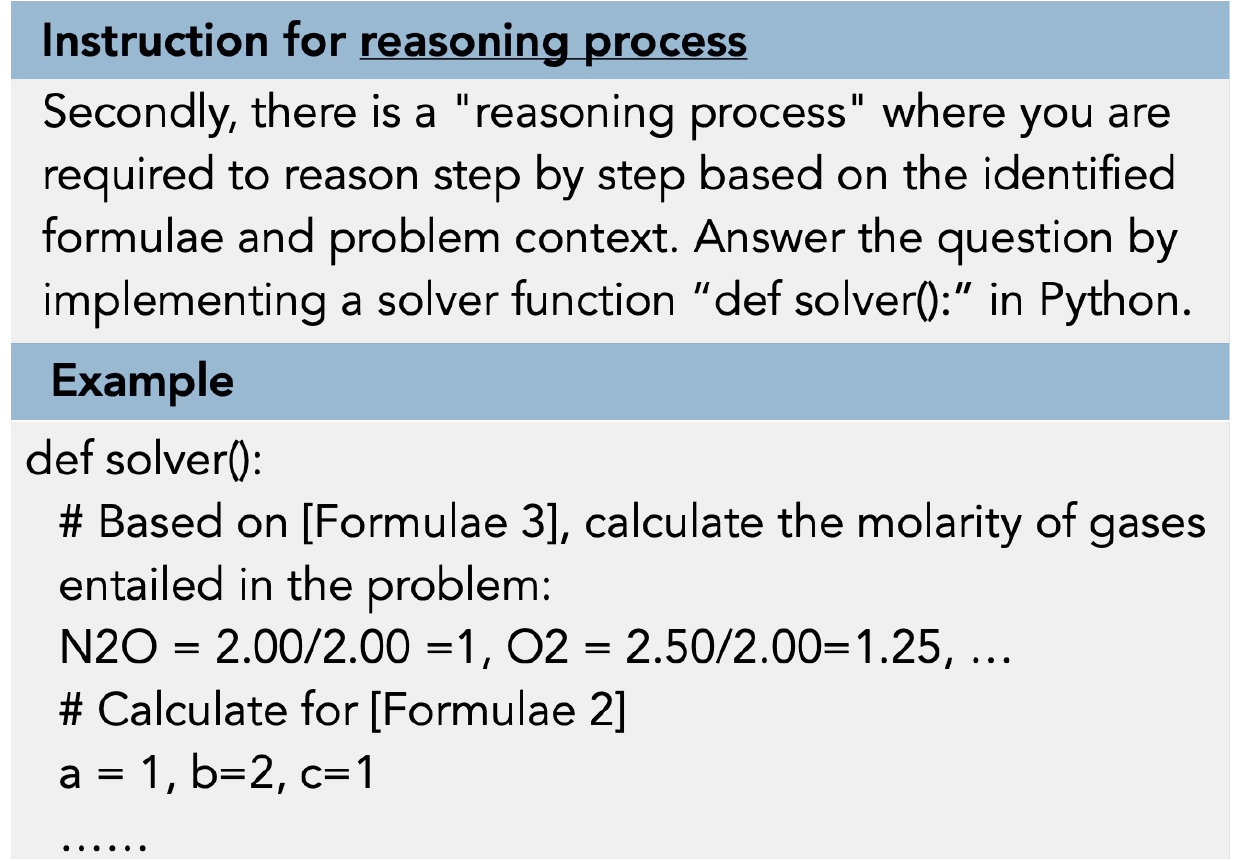}
    \caption{Instruction for reasoning process (part of the \textit{structured instruction} in Figure~\ref{fig:method}) and the output for problem in Figure~\ref{fig:intro}.}
    \label{fig:reasoning}
    % \vspace{-5mm}
\end{figure}

Grounded on the generated formulae, the LLMs can then reason about the solution to the original question. To induce LLMs for more precise reasoning and calculation processes, we adopt program-of-thoughts (PoT)~\cite{chen2023program} as demonstrated in Figure~\ref{fig:reasoning}. The detailed calculation process is translated into Python codes, accompanied by the annotation lines for reasoning. Concretely, we feed the problem and the \textit{structured instruction} into an LLM $\mathcal{M}$ as shown in the top of Figure~\ref{fig:method}. The generated results of this stage are formalized as $S_0 = (\{\mathcal{F}_0, C_0^f\}, \{\mathcal{R}_0, C_0^r\})$  where $\mathcal{F}_0=\{f_1,...,f_n\}$ denotes the formulae collected from Sec~\ref{sec: phase1}. $f_n$ is the $n$-th formula that is related. Similarly, $\mathcal{R}_0$ denotes the reasoning process. The output samples of $\mathcal{F}_0$ and $\mathcal{R}_0$ could be found in Figure~\ref{fig:formulae} and Figure~\ref{fig:reasoning}. $C_0^f$ and $C_0^r$ are the confidence scores for each part that is going to be mentioned Sec~\ref{sec:refinement}.

\subsection{Confidence-based Review-and-Refinement}
\label{sec:refinement}

The generated formulae and step-by-step reasoning are not always error-free. The cumulative errors in the formulae generation or step-by-step reasoning process can amplify and propagate throughout the entire generation, leading to wrong answers. Inspired by recent works of iterative prompting~\cite{wang-etal-2022-iteratively} and multi-model collaboration~\cite{DBLP:journals/corr/abs-2306-05685}, we employ the same LLM to conduct iteratively review-and-refine the previous iterations of generation. In each iteration, the LLM is instructed to revise the formulae and reasoning steps from the previous iteration.

During the review process, we found that there are chances when correct answers are swayed by incorrect ones after revision. This phenomenon echoes the recent findings~\cite{stechly2023gpt, huang2023large} questioning the correction ability of LLMs existing in prevalent works such as self-consistency~\cite{wang2023selfconsistency} and self-verification~\cite{weng2022large}.  To fix this, we estimate a confidence score $C_i$ on the revision process. Only a high-confidence revision is accepted for further refinement in the next iteration. The confidence assessment ensures each iteration makes meaningful progress towards final answers. 

Specifically, we prompt the LLM to provide confidence scores $C_i$ on a scale of $[0,1]$ for the $i$-th iteration. The review process starts with the generated formulae $\{\mathcal{F}_0, C_0^f\}$, where $C_0^f$ is the initial score for formulae generation. During the iterative review process, we leverage the same LLM to judge whether the collected formulae are correct with the confidence score $C_i^f$ of its own prediction. Formulae with the highest confidence score are kept. We then repeat it for the reasoning process so that we get the most faithful generations for both parts. In this way, LLMs can select the most aligned combination of problem, formulae, and the reasoning process. The elements above are finally input into the same LLM to get the final answer. The overall pipeline of this stage is shown in Algorithm~\ref{Alg: Iterative Review}, where the prompts $\{p_{rev}, p_{gen}\}$ denote the instruction for ``review'' and ``generation'' with refined results.

%% file: sections/experiment.tex
\section{Experiments}

\begin{table*}[t]

\centering\centering\setlength{\tabcolsep}{7.0pt}
\small
\caption{Results on the test sets for the four datasets, \textit{quan}, \textit{chemmc}, \textit{atkins}, \textit{matter}. We compare with baselines from two different settings, a zero-shot setting with no demonstrations and a few-shot setting with $3$ demonstrations. We compute the accuracy scores with the approximation detailed in Section 4.3. The best results for each setting are highlighted in \textbf{bold} and the second-best results are \underline{underlined}.}\label{table:res}
\begin{tabular}{lcccccccccc}
\toprule
\multirow{2}{*}{Methods} &
\multicolumn{4}{c}{\textbf{GPT-3.5}} &\multirow{2}{*}{Avg.} & \multicolumn{4}{c}{\textbf{GPT-4}}&\multirow{2}{*}{Avg.} \\
\cmidrule(lr){2-5}
\cmidrule(lr){7-10}
& \textit{quan} & \textit{chemmc} & \textit{atkins}& \textit{matter} && \textit{quan} & \textit{chemmc} & \textit{atkins}& \textit{matter}& \\
\midrule
\textbf{\textit{Zero-shot setting}}\\
% \midrule
Direct Reasoning& 5.88&\textbf{28.21}&\underline{8.41}&4.08&\textbf{11.65}&8.82&25.64&14.95&18.37&16.95\\
System Instruction & \textbf{8.82}&20.51&4.67&2.04&9.01&14.71&23.08&27.10&\underline{22.45}&21.84\\
CoT  &2.94&\underline{23.08}&6.54&\underline{10.20}&10.69&14.71&\textbf{43.59}&\underline{28.04}&20.41& \underline{26.69}\\
PoT &0.0&7.69&0.0&2.04&2.43&11.76&20.51&25.23&16.33&18.46\\
% Prompting w/ Python&&&&&&0.00&0.00&0.00&0.00&0.00\\
\colorbox{lightorange}{\ours} &\underline{5.88}&15.38&\textbf{9.35}&\textbf{12.24}&\underline{10.71}&\textbf{20.59}&\underline{38.46}&\textbf{31.78}&\textbf{24.49}&\textbf{30.11}\\
\midrule
\textbf{\textit{Few-shot setting}} \\
% \midrule
Direct Reasoning& 5.88&23.08&9.35&8.16&11.62&14.71&28.21&20.69&14.29&19.48\\
System Instruction &\underline{11.76}&15.38&5.61&4.08&9.21&17.65&30.77&15.87&12.24&19.13\\
CoT &8.82&20.51&8.41&6.12& 10.97&17.65&\underline{46.15}&\underline{21.05}&26.53&27.85\\
PoT &8.82&\underline{33.33}&\underline{13.08}&16.33&\underline{17.89}&\underline{38.24}&41.03&\underline{21.05}&\underline{28.57}&\underline{32.22}\\
% \colorbox{lightorange}{\ours} & \textbf{20.59} &\textbf{35.90}&\textbf{20.56}&\textbf{20.41}&\textbf{24.37}&\textbf{38.24}&\textbf{53.85}&\textbf{56.07}&\textbf{32.65} &\textbf{45.20}\\
\colorbox{lightorange}{\ours}&\textbf{32.35} &\textbf{43.59}&\textbf{26.17}&\textbf{24.49}&\textbf{31.66}&\textbf{41.18}&\textbf{58.97}&\textbf{59.81}&\textbf{30.67}&\textbf{47.64}\\
\bottomrule
\end{tabular}
% \vspace{-3mm}
\end{table*}

\subsection{Setup}

In our experiments, we use four datasets taken from SciBench~\cite{wang2023scibench}. The datasets cover a wide range of subfields including quantum chemistry, physical chemistry, kinetics, and matter, etc. The detailed distribution of subfields is shown in Figure~\ref{fig:subtopics}. The four datasets are manually collected from college-level chemistry textbooks, and are selected to be more challenging with free-response answers. Each of the datasets is divided into two parts, $\mathcal{P}_w$ and $\mathcal{P}_s$. Here $\mathcal{P}_w$ contains the majority number of problems that without solutions. Meanwhile, problems in $\mathcal{P}_s$ are coupled with solutions. The complexity of these datasets could also be proved by the average number of formulae entailed (around $2$) and the average reasoning steps (around $5$) generated by ~\ours needed to solve the problems (Detailed distribution information could be found in Appendix~\ref{sec:appendix2}.)
The detailed statistics are shown in Table~\ref{table:datasets}. 

\begin{table}[t]
\centering\setlength{\tabcolsep}{4.5pt}
\small
  \caption{Detailed statistics and information of the four datasets we experiment with. $\#\mathcal{P}_w$ and $\#\mathcal{P_s}$ refer to the number of data samples with and without solutions. ``\# F'' means the average number of formulae entailed in the problem, and ``\# RS'' denotes the average reasoning steps for each problem.}
  \label{table:datasets}
  \begin{tabular}{lcccc}
    \toprule
    \textbf{Datasets}&\textbf{Subfields/Topics}&\textbf{\# $\mathcal{P}_w$($\mathcal{P}_s$)}&\textbf{\# F} &\textbf{\# RS}\\
    \midrule
    \textit{quan}  &Quantum chemistry&34 (8)&1.93&3.94\\
    \textit{chemmc} &Quantum mechanics&39 (9)&1.88&3.95\\
    \textit{atkins} &Physical chemistry&107 (16)&1.65&4.33\\
    \textit{matter}&Chemistry kinetics&49 (10)&1.89&4.43\\
  \bottomrule
\end{tabular}
% \vspace{-3mm}
\end{table}

% \begin{table}[t]
% \centering\setlength{\tabcolsep}{1pt}
% \small
%   \caption{Detailed statistics and information of the four datasets we experiment with.}
%   \label{table:datasets}
%   \begin{tabular}{lcccc}
%     \toprule
%     \textbf{Datasets}&\textbf{quan}&\textbf{chemmc}&\textbf{atkins} &\textbf{matter}\\
%     \midrule
%     \textit{Subfields}  &Quantum chemistry&Quantum mechanics&Physical chemistry&Chemistry kinetics\\
%     \textit{\# Problems} &42&48&123&59\\
%     \textit{\# Formulae} &1.93&1.67&1.65&1.89\\
%     \textit{\# Reasoning steps} &3.91&4.51&5.12&4.57\\
%   \bottomrule
% \end{tabular}
% \end{table}

% \subsection{Baselines and Settings}

Experiments are conducted under both zero-shot and few-shot settings. For the few-shot setting, the demonstrations are constructed with $3$ examples randomly sampled from $\mathcal{P}_s$. We leverage GPT-3.5 (gpt-3.5-turbo) and GPT-4 (gpt-4-0315) as our backbone models. For each setting, we consider four baselines following the evaluation paradigm in SciBench (Full instructions are provided in Appendix~\ref{sec:appendix1}): 

(i) \textbf{Direct reasoning} refers to directly feeding the problem into the model without any other instructions; 

(ii) \textbf{System instruction} is originally developed by~\citet{wang2023scibench} which is tailored to the task and describes the types and categories of questions, along with instructions; 

(iii) \textbf{CoT} follows the ``step-by-step'' prompting strategy that requires the model to output the ``thinking process'' first; 

(iv) \textbf{PoT} leverages the idea of Program-of-Thoughts (PoT)~\cite{chen2023program} which translates the solution into Python codes to improve the understanding and calculation ability of LLMs.

\subsection{Implementation Details}

We access the two LLMs, GPT-3.5 and GPT-4, with the OpenAI API. During our experiments, the temperature for generation is kept at $0$ to ensure reproducibility and reduce potential variances. When doing the evaluation, we follow the previous work~\cite{wang2023scibench} and compare the model outputs with the correct answers, allowing an absolute deviation of $0.1$ for answers greater than $1$ and a relative tolerance of $0.05$ for answers less than $1$. This guarantees a fair comparison with previous baseline models.

\subsection{Results}

Table~\ref{table:res} presents the performance of all methods on the test set of the four datasets. We report the model performance in terms of accuracy scores. The best results are bolded. Based on the results, we have the following key observations:

(i) \textbf{\ours achieves superior performance on almost all the datasets in both zero-shot and few-shot settings.} Specifically, \ours achieves an absolute improvement of $+13.77$ and $+15.42$ in terms of the average score on few-shot settings, respectively, which is $43.49\%$, and $32.37\%$ of relative improvement. The notable improvement demonstrates the effectiveness of our model in adapting to various scenarios by inducing chemistry knowledge and performing precise reasoning. 

(ii) \textbf{\ours has shown effectiveness generally across different backbones.} In addition, using GPT-4 as backbone models consistently outperforms GPT-3.5 backbones for all methods experimented by a large margin. Specifically, we found that \ours achieves more pronounced performance improvement with GPT-4, with $+2.43$ and $+6.50$, respectively. Given the fact that GPT-4 is a more powerful model than GPT-3.5, this result is encouraging, as it suggests that as foundational models continue to evolve, \ours can be expected to provide even greater benefits. 

(iii) \textbf{\ours delivers greater performance improvements on four datasets in few-shot settings compared to zero-shot settings.} With GPT-4, the performance improvement brought by \ours is $+12.0$ more than the zero-shot setting. The reason could be twofold. One is the complexity of these datasets, because it is hard to directly output answers or solutions without any demonstrations as references. Another factor could be the format of solutions offered by \ours, which disentangles the solution with structured instructions. This format is hard to learn with no examples. Contrastively, we do not observe obvious differences in performance improvement brought by baselines for few-shot and zero-shot settings. This further shows \ours's ability to learn by analogy.

(iv) \textbf{\ours achieves substantial performance gains in complex problems with extensive reasoning steps.} Specifically, we found the performance on datasets \textit{chemmc} and \textit{atkins} is better than \textit{quan} and \textit{matter} with GPT-4 in few-shot setting. \ours performs particularly well on \textit{atkins} dataset, with the accuracy scores doubling in almost all settings. We attribute the reason to the number of formulae in \textit{atkins}. As shown in Table~\ref{table:datasets}, \textit{atkins} has the smallest average number of formulae, making it easier for formulae collection. However, it has a larger number of average reasoning steps. This verifies \ours's ability to deal with complicated reasoning processes.

%% file: sections/analysis.tex
\section{Analysis}\label{sec:analysis}

Intuitively, we want to validate the quality of produced chemistry reasoning processes. We first fine-tune smaller models using generated reasoning outputs. We then conduct a thorough ablation study of \ours' various components to gain a deeper understanding of its effectiveness. Additionally, an error analysis further offers insights about how to make \ours even better in the future. 
% We use GPT-4 as the backbone model across the analyses.
 % to reduce influencing factors and avoid variances. 

\subsection{Validating the Reasoning Quality}

Our method \ours has shown strong improvement in accuracy over baselines. 
Here we further validate that \ours generates high-quality intermediate reasoning steps that increase answer accuracy.
Specifically, we fine-tune smaller language models, Llama-2-13B-chat~\cite{touvron2023llama2} and Vicuna-13B~\cite{vicuna2023}, on the reasoning steps generated by \ours and CoT, respectively. The rationale is that while a smaller model may already be equipped with some domain knowledge, it typically lacks the capability for step-by-step reasoning in complex chemistry problems---a skill that emerges predominantly in larger-scale models. By fine-tuning smaller models with the generated reasoning steps, we essentially teach them to perform this advanced reasoning. Intuitively, using higher-quality fine-tuning data would lead to better performance in the small models.

% we are also curious about how it will improve the problem-solving ability on smaller open-source language models.

% Following previous practices~\cite{magister-etal-2023-teaching, liang2023let}, We further fine-tune open-source LLM, Vicuna-13B~\cite{vicuna2023}, with generation results of \ours on $\mathcal{P}_w$, which include all the contents generated introduced in Section~\ref{sec:methdology}. 

% Data samples in $\mathcal{P}_s$ are used for testing.
To \textit{collect fine-tuning data}, we first instruct GPT-4 to generate another $1,000$ problems with $3$ problems sampled from SciBench as demonstrations. To encourage diversity, we set the generation temperature as $1.0$ and filter out problems that have $5$-gram or larger overlapping with existing generated problems. Then, we use \ours for providing the solutions to all $1,000$ problems as the paired training data.
Additionally, we compare two other baselines: (i) Fine-tuning model on the original data, which only consists of the original problem statement and the direct answer, formatted as ``[problem] The answer is therefore [answer]''; (ii) Pure zero-shot inference, where given the problem as input, the model outputs a direct answer without any fine-tuning.

% with the following two baselines: (i) Vanilla training data that is only composed of the original problem statement and the direct answer prepended with ``The answer is therefore [a]'', and (ii) Training data that incorporates reasoning process generated by CoT prompting together with the problem context. 
 The fine-tuning process is based on LoRA~\cite{hu2022lora}, a parameter-efficient fine-tuning method.
 % , and we mask out the input for loss computation. 
For details on training and problem generation, please refer to Appendix~\ref{sec:appendix3}.

Results are shown in Table~\ref{table:fine-tune}. The vanilla version falls short in solving such complex chemistry problems, as shown by their zero-shot performance on four datasets. Training only on the original problem and answer pairs does not bring much improvement compared with direct inference. Fine-tuning with data generated by \ours, on the other hand, brings more than $20\%$ absolute improvement. % It further shows the potential of \ours to teach smaller language models for complex chemistry problems.
Fine-tuning based on \ours is superior to fine-tuning with CoT, demonstrating that \ours can produce detailed reasoning at a higher quality.

\begin{table*}[t]

\centering\centering\setlength{\tabcolsep}{6.5pt}
\small
\caption{Ablation studies for different components in \ours in both zero-shot and few-shot settings with the backbone model GPT-4. The accuracy scores are reported with all four datasets.}\label{table:ablation}
\begin{tabular}{lcccccccccc}
\toprule
\multirow{2}{*}{Methods} &
\multicolumn{4}{c}{Zero-shot} &\multirow{2}{*}{Avg.} & \multicolumn{4}{c}{Few-shot}&\multirow{2}{*}{Avg.} \\
\cmidrule(lr){2-5}
\cmidrule(lr){7-10}
& \textit{quan} & \textit{chemmc} & \textit{atkins}& \textit{matter} && \textit{quan} & \textit{chemmc} & \textit{atkins}& \textit{matter}& \\
\midrule
structured instruction &23.53&35.90&39.25&18.37&29.26&32.35&51.28&53.27&28.57& 41.37 \\
\quad + review for $\mathcal{F}$ &\underline{26.47}&38.46&40.19&18.37&30.87&32.35&48.71&54.55& 30.61&41.56 \\
\quad + review for $\mathcal{R}$ &23.53&41.03&\underline{41.12}&22.45&\underline{32.03}&35.29&51.28&54.55&30.61&42.93 \\
\quad + confidence score&\textbf{29.41}&\textbf{41.03}&\textbf{46.34}&\underline{23.08}&\textbf{34.97}&38.24&53.85&56.07&32.65 &45.20\\
\quad + PoT &20.59&\underline{38.46}&31.78&\textbf{24.49}&30.11&\textbf{41.18}&\textbf{58.97}&\textbf{59.81}&\textbf{30.67} &\textbf{47.64}\\
% \textbf{\ours}&\textbf{-}&\textbf{-}&\textbf{-}&\textbf{-}&\textbf{-}&\textbf{41.18}&\textbf{58.97}&\textbf{59.81}&\textbf{30.67} &\textbf{47.64}\\
% \midrule
% - PoT &29.41&41.03&46.34&23.08&34.97&38.24&53.85&56.07&32.65 &45.20 \\
% - confidence score&23.53&41.03&41.12&22.45&32.03&35.29&51.28&54.55&30.61&42.93 \\
% - iterative review&23.53&35.90&39.25&18.37&29.26&32.35&51.28&53.27&28.57& 41.37\\
% - review for $\mathcal{F}$&26.47&38.46&40.19&18.37&30.87&32.35&48.71&54.55& 30.61&41.56\\
% \midrule
% w/ PoT &-&-&-&-&-&41.18&58.97&59.81&30.61&47.64 \\
\bottomrule
\end{tabular}
% \vspace{-3mm}
\end{table*}

\begin{figure}[t]
    \centering
    \includegraphics[width=0.95\linewidth]{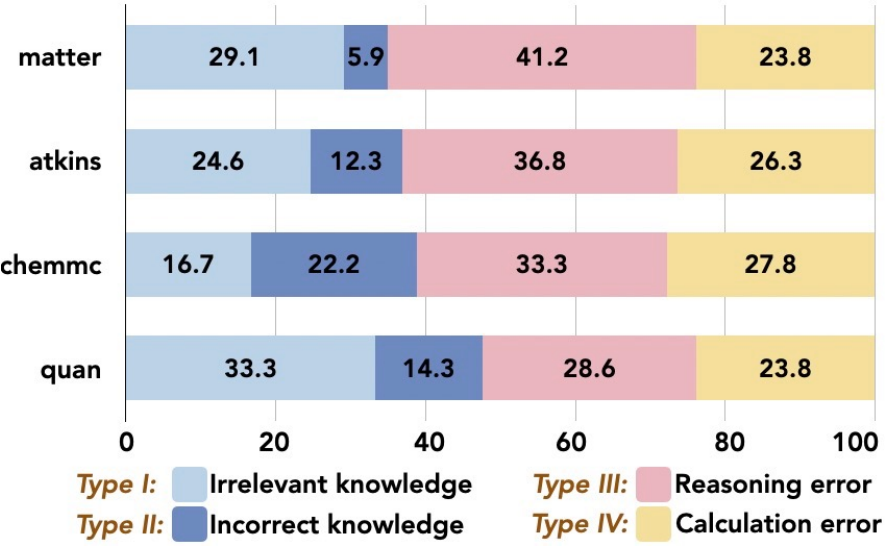}
    \caption{Error analysis of four error categories across all datasets (y-axis) in terms of error proportions (x-axis). The results are for \ours \textit{w/o PoT} to exclude the influence of external tools.}
    \label{fig:error}
    % \vspace{-3mm}
\end{figure}

\begin{table*}[t]

\centering\centering\setlength{\tabcolsep}{7.0pt}
\small
\caption{Fine-tuning results with generations from \ours as training data on two open-source models. The accuracy scores are reported with all four datasets.}\label{table:fine-tune}
\begin{tabular}{lcccccccccc}
\toprule
\multirow{2}{*}{Methods} &
\multicolumn{4}{c}{Llama-2-13B-chat} &\multirow{2}{*}{Avg.} & \multicolumn{4}{c}{Vicuna-13B}&\multirow{2}{*}{Avg.} \\
\cmidrule(lr){2-5}
\cmidrule(lr){7-10}
& \textit{quan} & \textit{chemmc} & \textit{atkins}& \textit{matter} && \textit{quan} & \textit{chemmc} & \textit{atkins}& \textit{matter}& \\
\midrule
\textbf{Zero-shot inference}&0.0&0.0&0.0&2.04&0.51&5.88&2.56&3.74&0.0&3.05\\
\midrule
Original&0.0&5.13&0.0&0.0&1.28&8.82&5.13&2.80&2.04&4.70\\
CoT &8.82&17.95&9.35&8.16&11.07&11.76&15.38&8.41&6.12&10.42\\
\colorbox{lightorange}{\ours}&14.71&30.77&20.56&20.41&21.61&17.65&33.33&18.69&20.41&22.52\\
\bottomrule
\end{tabular}
% \vspace{-3mm}
\end{table*}

\subsection{Ablation Study}

To understand why \ours works particularly well, we ablate \ours with add-ons from different components. Table~\ref{table:ablation} summarizes the experimental results. 

Firstly, both ``structured instruction'' and ``iterative review and refinement'' are significant in contributing to the performance of \ours for zero-shot and few-shot settings. Specifically, removing the confidence score and iterative review resulted in a decrease of $2.27$ and $3.83$, respectively.

It is worth noting that while iterative refinement indeed contributes to the performance, our strategy of structured instruction is strong enough and demonstrates comparative performance with strong baselines such as CoT. When removing iterative review for formulae $\mathcal{F}$ alone, the performance drops by a large margin, which is comparable to removing the whole iterative review process. This shows the effectiveness of iterative review for formulae collection and the importance of domain knowledge when solving. 

Also, though PoT helps with precise calculation and improves performance, \ours without PoT still outperforms the strongest baselines. We also note that in the zero-shot setting, \ours \textit{without PoT} achieves even stronger performance. This may be attributed to the decrease in the instruction-following ability of LLM for codes when there is no demonstration provided.

\subsection{Error Analysis}\label{sec:error_analysis}

We conduct a manual analysis of all the $113$ error cases for \ours \textit{without PoT} with GPT-4 as the backbone for few-shot setting across four datasets. Error types are defined as corresponding to the two processes of ``formulae generation'' and ``step-by-step reasoning''. The analysis is done by $3$ Ph.D. students with a chemistry background.

For ``formulae generation'', we define two types of errors that are related to this process. \textbf{Irrelevant knowledge} indicates that the formulae collected are not relevant to solving the problem. For example, solving a problem requires \textit{Broglie formula} but LLM collects \textit{Wavelength formula}. \textbf{Incorrect knowledge} refers to the incorrectness inherent in the formula itself. $K_c=\frac{[N_2O]}{[N_2]\times[O_2]}$ in Figure~\ref{fig:intro} is one such example. For ``step-by-step reasoning'', we also have two error cases as follows. \textbf{Reasoning error} refers to the errors made during the intermediate reasoning steps. For example, in Figure~\ref{fig:intro}, the model fails to reason the correct relations of different gases during the reaction $O_2 + N_2 \rightarrow N_2O$. \textbf{Calculation error} means the mathematical computation mistakes made when doing the reasoning process. 

The results are shown in Figure~\ref{fig:error}, where we plot the proportion for every error type of each dataset. We have the following key observations:

(i) \textbf{\ours are more likely to generate irrelevant formulae than inaccurate ones.} On average, only $13.7\%$ of the total errors are caused by incorrect forms of formulae compared with an average of $25.9\%$ of irrelevant ones. The irrelevance rate is slightly higher than that of GPT-4 (CoT) as shown in Figure~\ref{fig:intro-error}. A potential reason is that \ours could focus on the irrelevant formulae collected in the first phase. For the entire formulae collection process, although \ours sometimes retrieve irrelevant formulae for solving a problem, the formulae are less likely to be incorrect themselves. 

(ii) \textbf{Formulae being relevant probably is more important than being correct.} We observed that the ``Irrelevance'' rate is relatively low for \textit{atkins} and \textit{chemmc} datasets, although they may have higher ``incorrectness'' rate. This is potentially another explanation of why performance in Table~\ref{table:res} is particularly high for these two datasets compared with the rest two datasets, since the formulae collection process serves as a necessary condition for conducting correct reasoning processes further. 

(iii) \textbf{Complex reasoning ability is still the bottleneck of LLMs.} Although \ours drastically decreases the reasoning error as shown in Figure~\ref{fig:intro-error}, ``reasoning error'' still takes up to around $35.0\%$ of all error cases. For chemistry problems that entail multiple elements interacting in a complex environment, the ability to reason out the relations among objects becomes crucial.

(iv) \textbf{Preciseness is important for solving complex chemistry problems.} Without PoT, the case of ``calculation error'' still occupies a large portion of around a quarter. Even a single step of calculation error could lead to wrong answers in chemistry reasoning problems. 

\begin{figure}[t]
    \centering
    \includegraphics[width=0.95\linewidth]{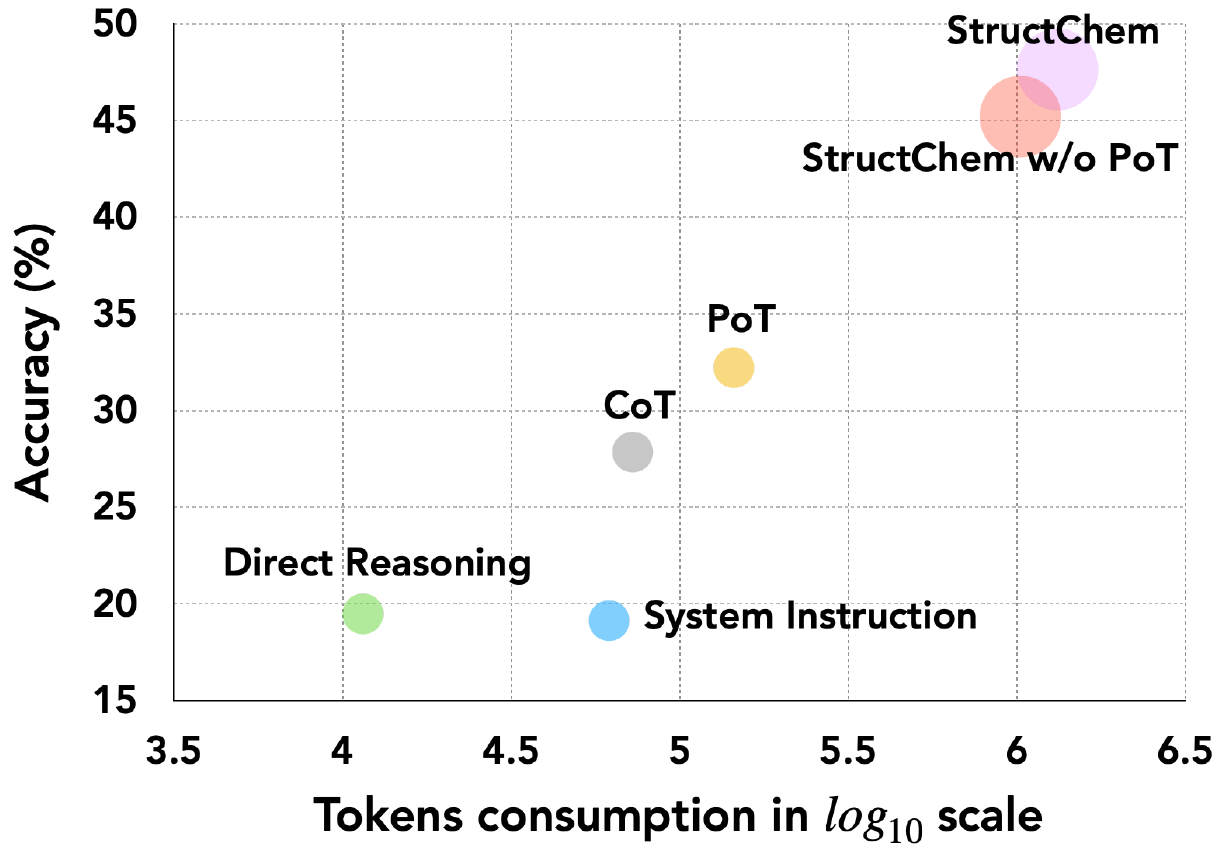}
    \caption{Cost-effectiveness analysis. The size of each dot is proportional to the average number of inferences by each method. The y-axis denotes the average accuracy across four datasets.}
    \label{fig:cost}
    % \vspace{-5mm}
\end{figure}

\subsection{Cost-Effectiveness Analysis}

By introducing \ours, we manage to reduce the costs associated with complex chemistry problems while achieving comparable or even superior performance. We conduct experiments in the few-shot setting with GPT-4 as the backbone. We define cost as the sum of tokens for instruction, demonstrations, and output. Based on results illustrated in Figure~\ref{fig:cost}, we can see that the performance increase brought by \ours is actually a little larger compared to CoT and PoT considering the ratio of tokens consumption. \ours's substantial improvement does not rely on the consumption of tokens.

%% file: appendix.tex
\section{Prompts used for baseline methods in Section 4.}
\label{sec:appendix1}

In this section, we provide the detailed prompts used for experiments.

\begin{figure}[h]
% \vspace{-3mm}
    \centering
    \includegraphics[width=0.9\linewidth]{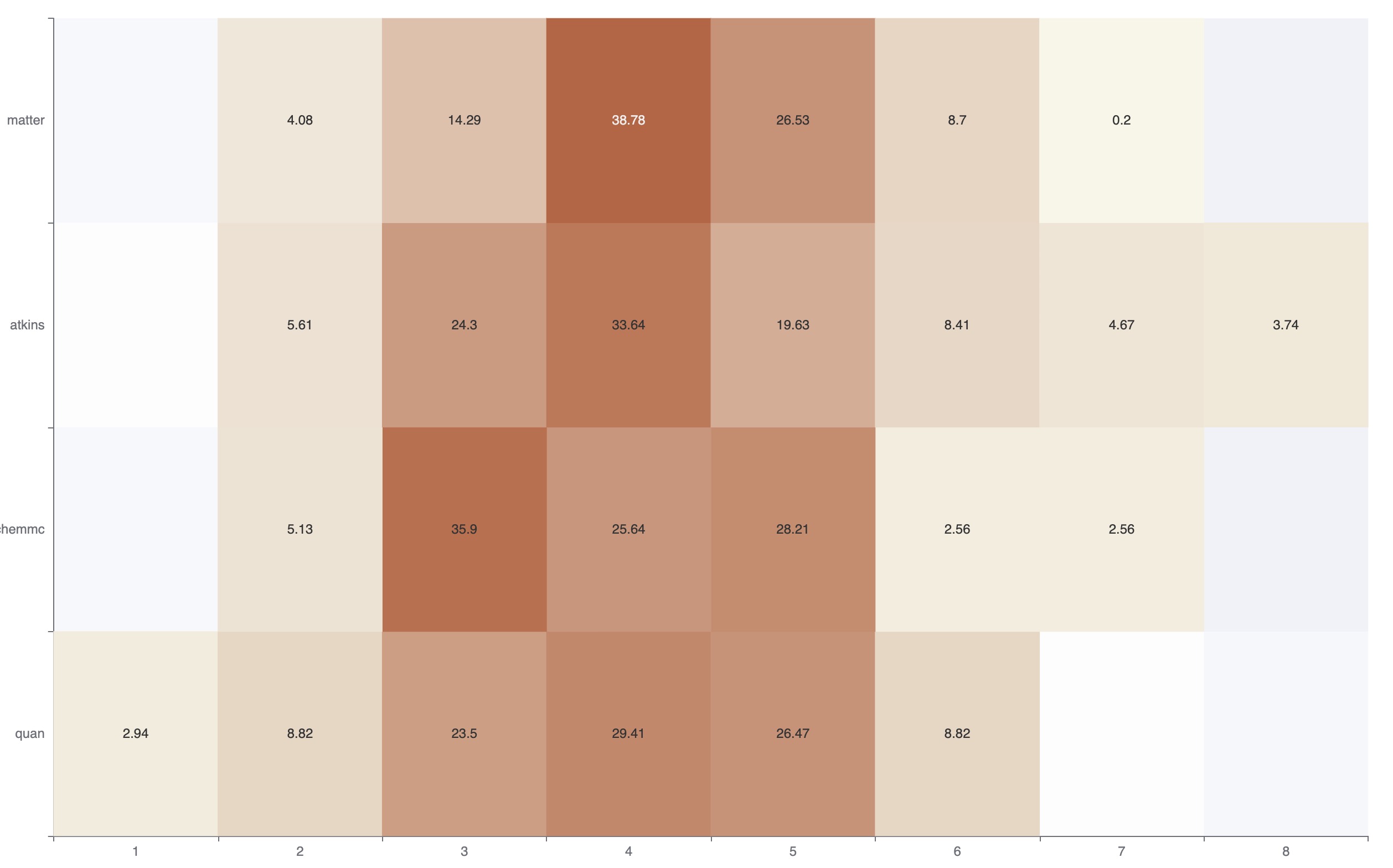}
    \caption{Distribution of four datasets in terms of the reasoning steps.}
    \label{fig:distribution}
    \vspace{-3mm}
\end{figure}

\paragraph{System Prompt} Please provide a clear and step-by-step solution for a scientific problem in the category of Chemistry. The problem will specify the unit of measurement, which should not be included in the answer. Express the final answer as a decimal number with three digits after the decimal point. Conclude the answer by stating ``The answer is therefore [ANSWER].''

\paragraph{Program-of-Thought Prompt} Please provide a clear and step-by-step solution for a scientific problem in the category of Chemistry. The problem will specify the unit of measurement. Please translate the solution steps into Python code and encase the Python code within triple backticks for clarity.

\begin{figure*}[t]
    \centering
\includegraphics[width=0.96\linewidth]{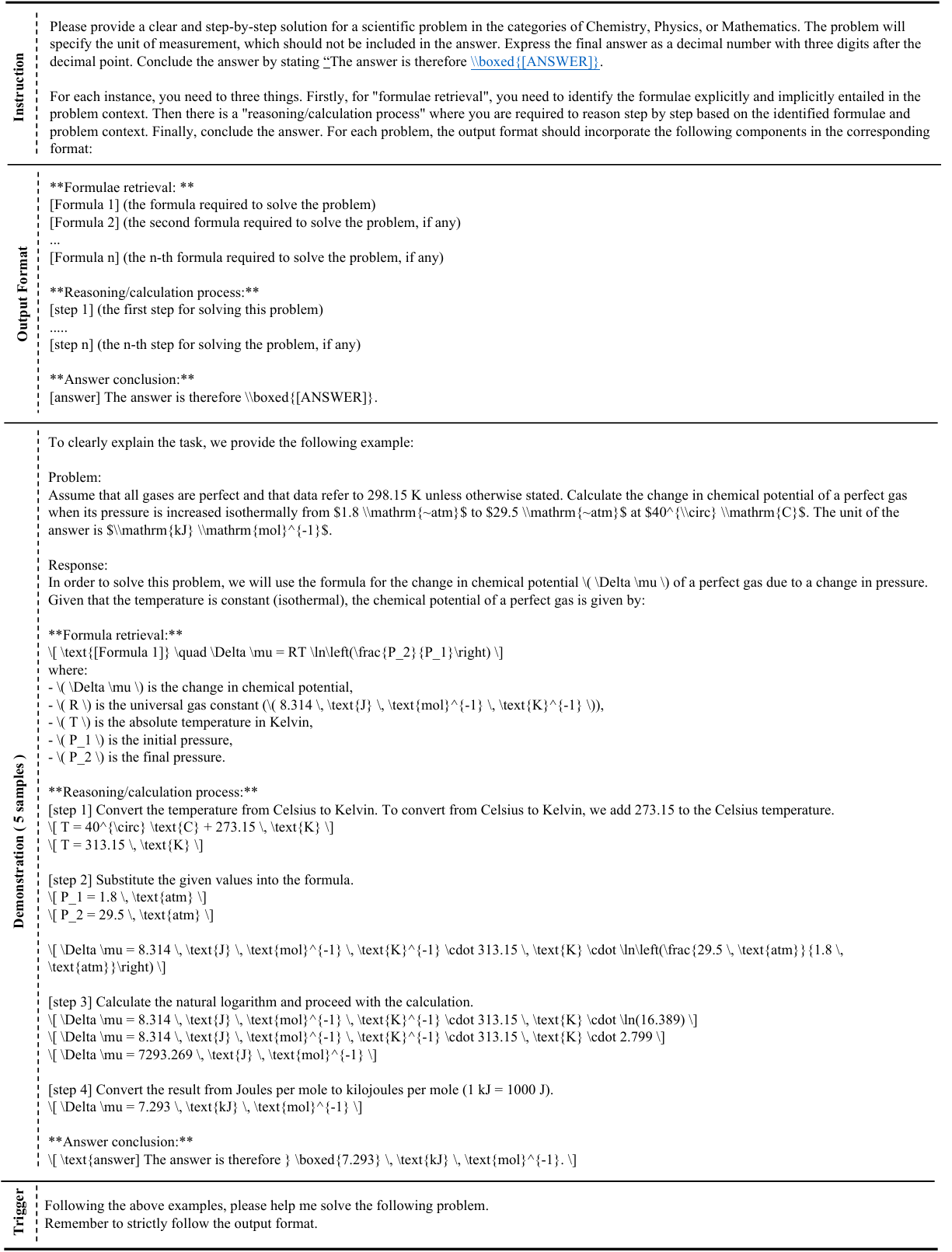}
    \caption{Full prompt used for generation.}
    \label{fig:prompt}
\end{figure*}

\paragraph{Template-guided Prompt} The full prompt for ``formulae generation'' and ``step-by-step reasoning'' is composed of four stages, general instruction, output format, demonstrations, and trigger. The complete view of the prompt is shown in Figure~\ref{fig:prompt}.

\section{Distribution of datasets}\label{sec:appendix2}

The detailed distribution of four datasets in terms of reasoning steps is shown in Fig~\ref{fig:distribution}. We can see that the majority of the samples have reasoning steps spanning $[3,5]$. Some samples even have reasoning steps of $8$, which demonstrate the complexity of these datasets.
\begin{figure}[t]
    \centering
    \includegraphics[width=0.8\linewidth]{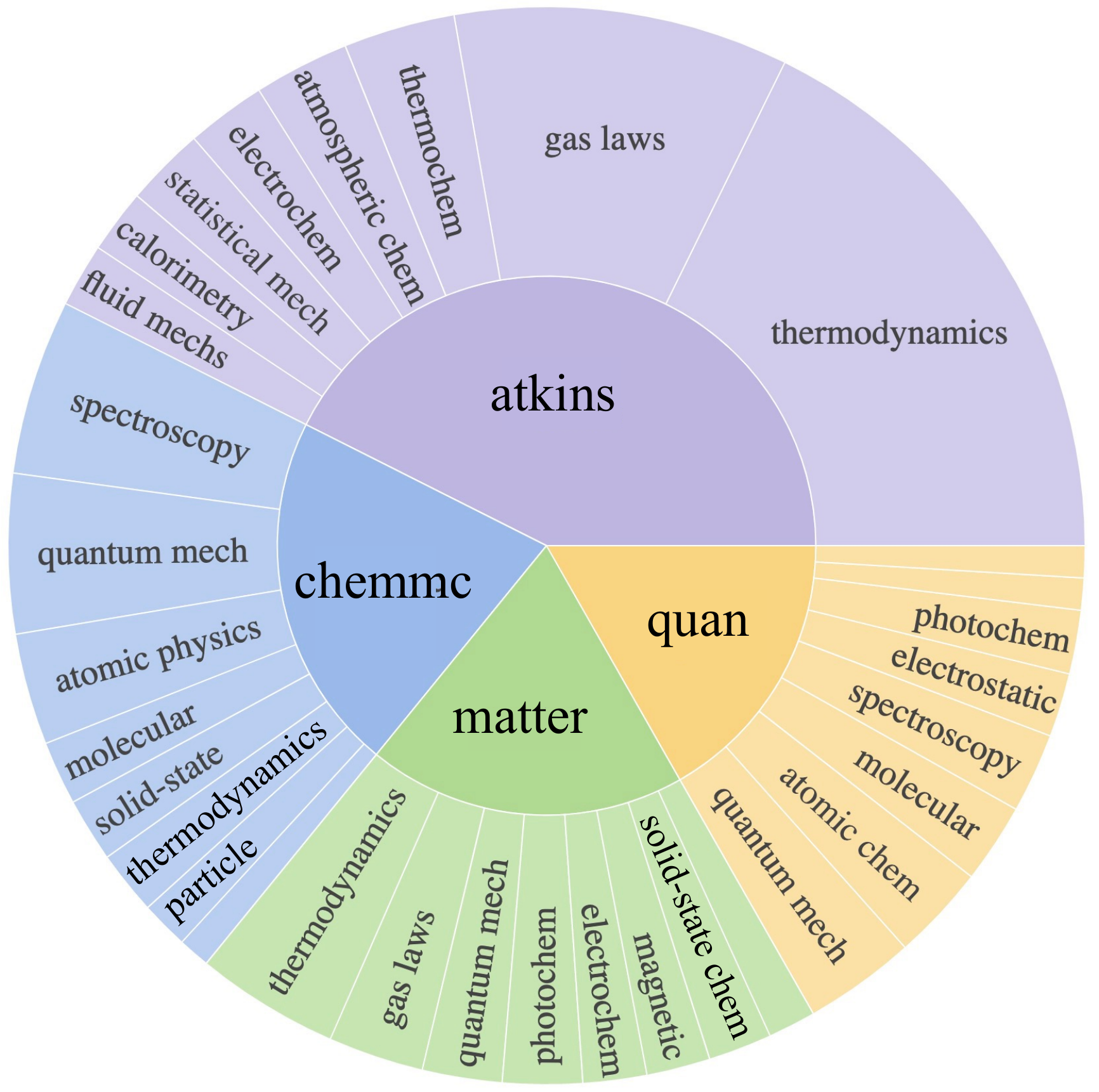}
    \caption{\colorbox{orange1}{\textit{Quantum chemistry (quan)}}~\cite{hairpearson} provides an exploration of equilibrium, structure, and reactions. \colorbox{green}{\textit{Chemistry kinetics (matter)}}~\cite{atkins2014physical} combines physics and mathematics, spanning through quantum mechanics and atomic structure. \colorbox{blue}{\textit{Quantum mechanics (chemmc)}}~\cite{mcquarrie2008quantum} covers quantum mechanics and the applications in chemical bonding. \colorbox{purple}{Physical chemistry (atkins)}~\cite{atkins2023atkins} provides explorations of equilibrium, structure, and reactions. We leverage GPT-4 to annotate each data sample in these datasets for the specific subfields.}
    \label{fig:subtopics}
    \vspace{-3mm}
\end{figure}

\begin{figure*}[t]
    \centering
    \includegraphics[width=0.9\linewidth]{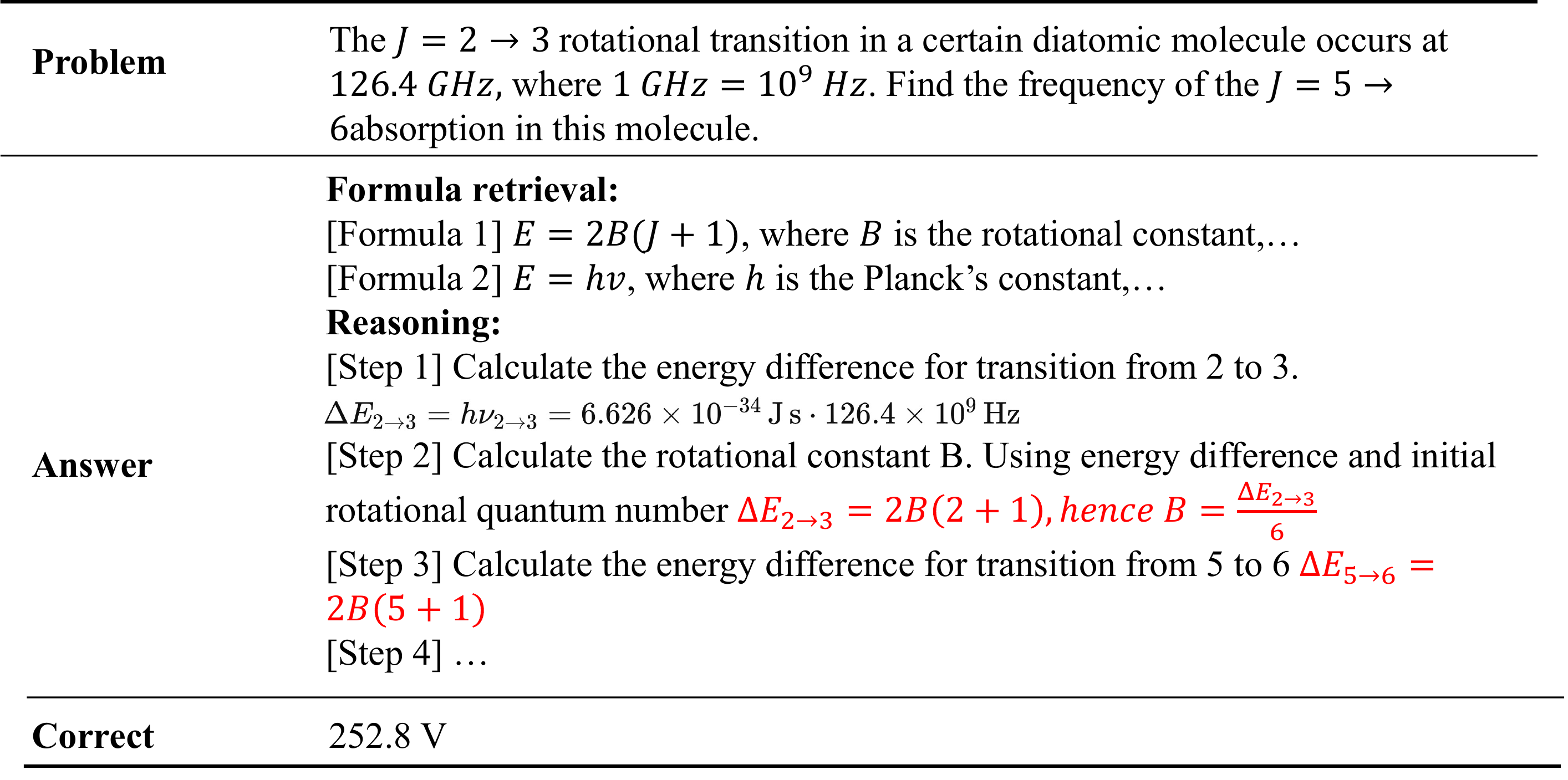}
    \caption{An example of a reasoning error. The red highlighted expression is deduced from the given formula. Instead of solving for difference, it directly plug the value for calculating energy. The correct expression for the first one should be $\Delta E_{2\rightarrow 3}=2B(3+1)-2B(2+1)$.}
    \label{fig:reasoning_error}
\end{figure*}

\begin{figure*}[t]
    \centering
    \includegraphics[width=0.9\linewidth]{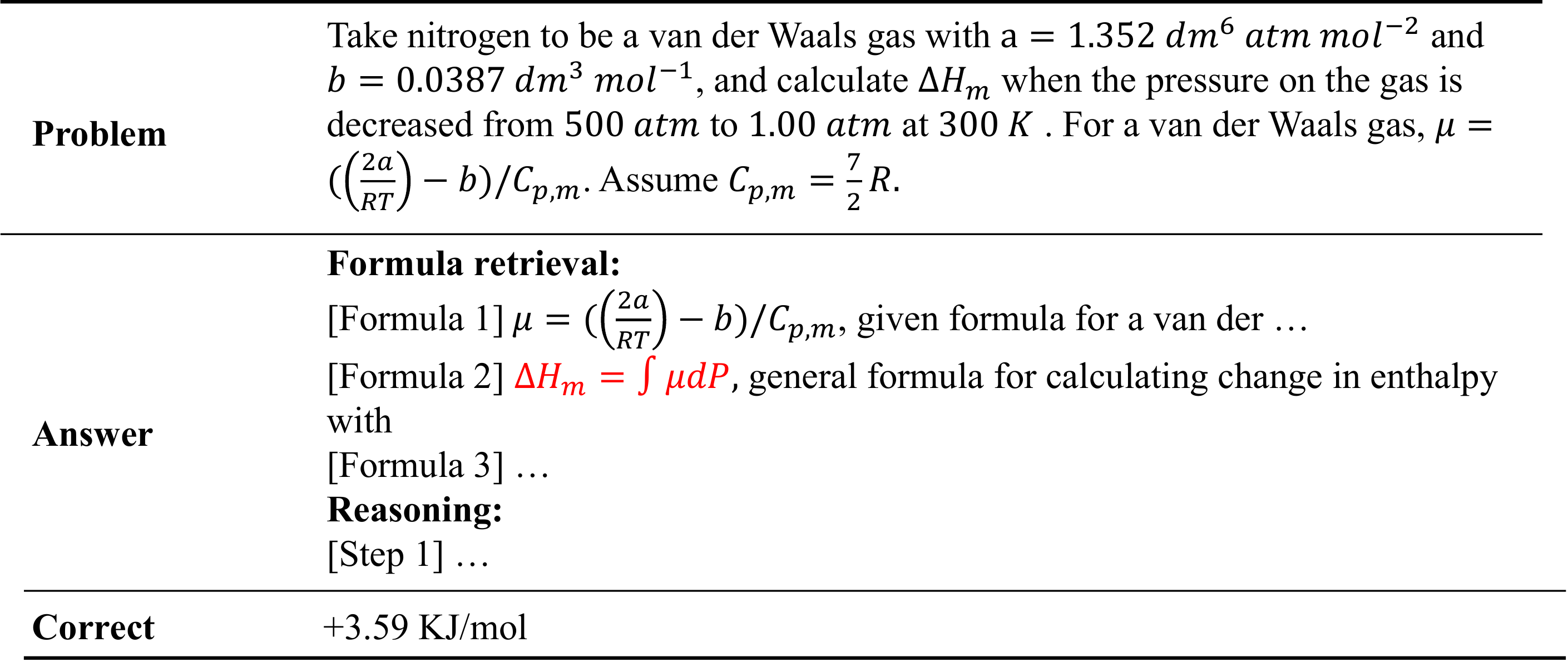}
    \caption{An example of a factual error. The red highlighted expression is the wrong form to calculate. The correct formulae should be $\int_{p}^{p'}dH_m$, which could be further converted to $H_mqx=H_m(v'-d)$.}
    \label{fig:fac_error}
\end{figure*}

\section{Details for Section 5.3}\label{sec:appendix3}

\paragraph{Instruction for problem generation} Please help me to generate complex and difficult chemistry problems that include but are not limited to the fields of physical chemistry, quantum chemistry, thermodynamics, atomic chemistry, molecular, etc. To help you better understand, I provide the following examples: [demonstrations]. Following the above examples, please help me with this task and generate three problems that satisfy my requirements. Make sure the generated problems are reasonable and complex for solving.

\paragraph{Training details}
We use LLaMA-2-13B-chat~\cite{touvron2023llama2} and Vicuna-13B-v1.3~\cite{vicuna2023} as backbone models and finetune them with the LoRA approach~\cite{hu2022lora}.
During training, we configure the batch size to $8$ and the maximum learning rate to 1e-4 with a 0.03 warmup ratio.
For all the experiments, the LoRA $r$ is set to 8, and we apply a dropout rate of 0.05.
We keep these hyperparameters the same for a fair comparison. We train the models with $10$ epochs and it takes around $1$ hour to train on a single NVIDIA A6000 GPU. 
During the inference process, we also adhere to the same set of parameters: a temperature of 0.1, top\_p of 0.75, top\_k of 40, 4 beams, and a maximum generation length of 2,048.

\section{Examples of error type in Section 5.2}

\begin{figure*}[t]
    \centering
    \includegraphics[width=0.9\linewidth]{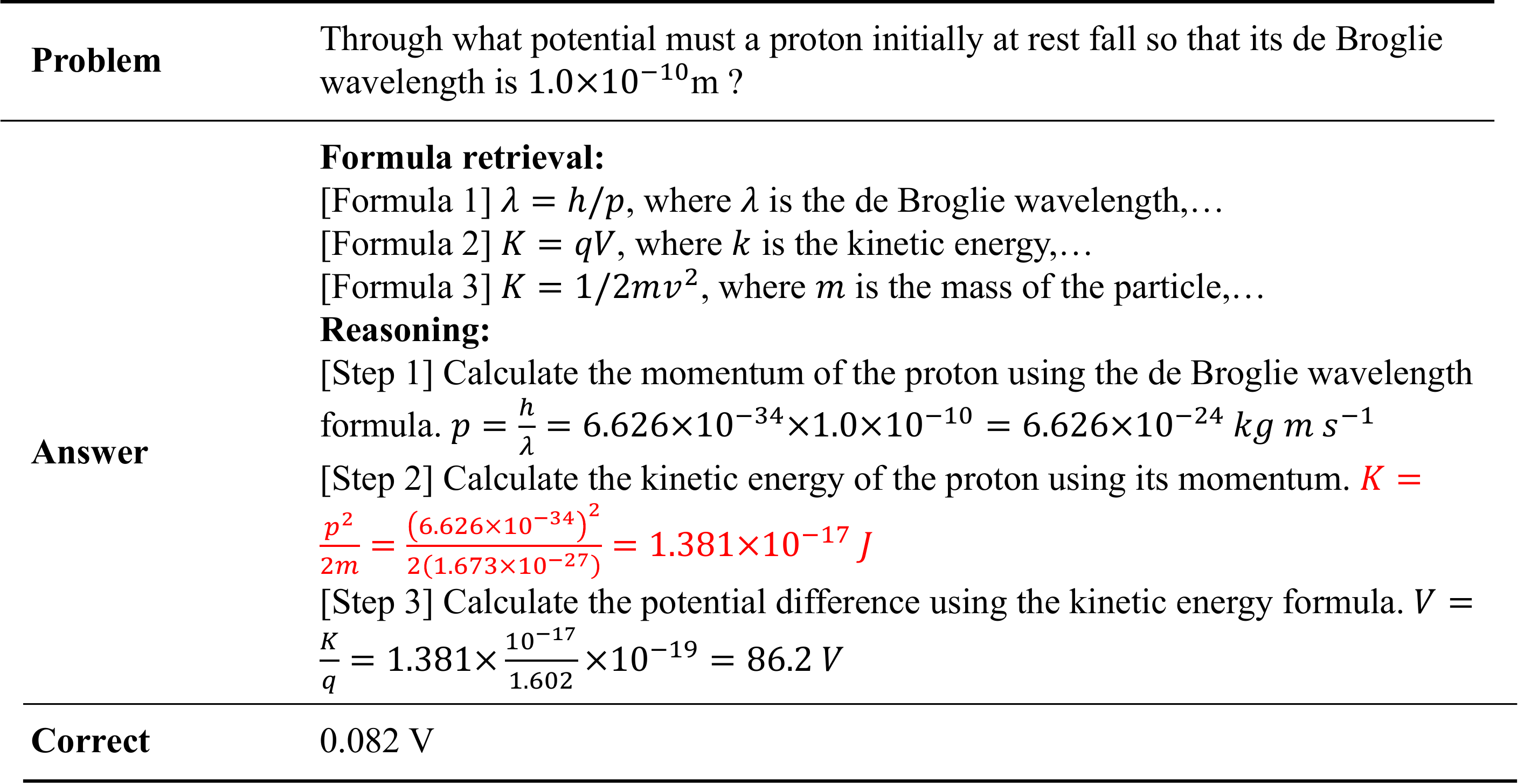}
    \caption{An example of a calculation error. The red highlighted expression is wrongly calculated. If calculated correctly, we will get $K=1.312\times 10^{-20}$, which yields the correct answer. }
    \label{fig:cal_error}
\end{figure*}

\begin{figure*}[t]
    \centering
    \includegraphics[width=0.9\linewidth]{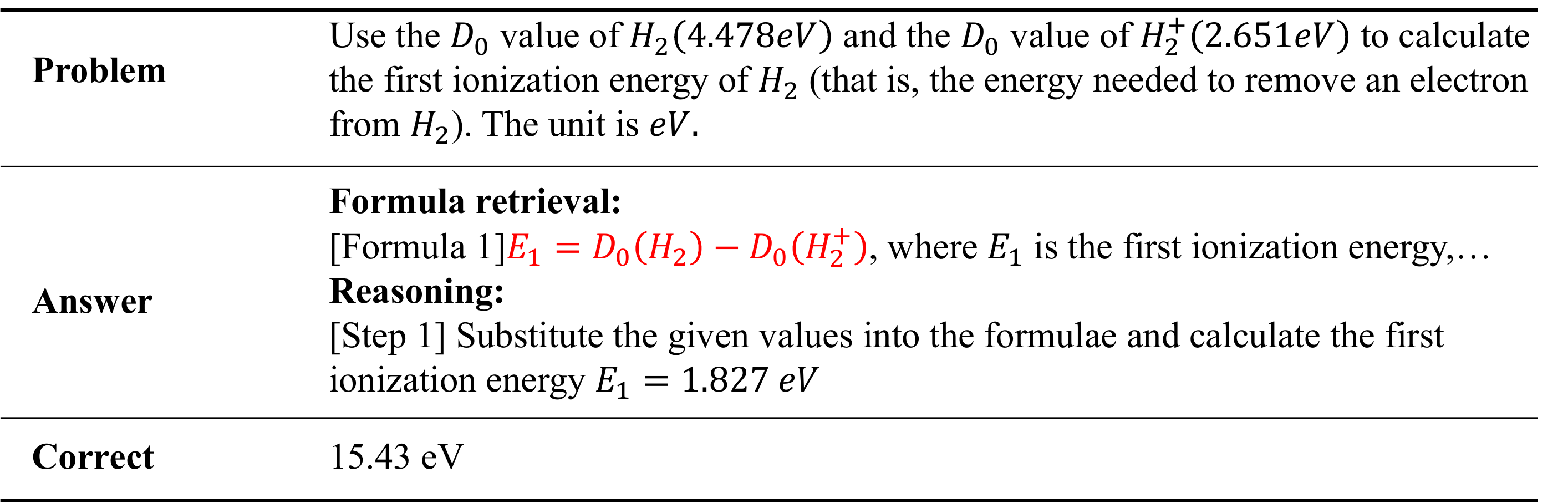}
    \caption{An example of a principle error. The red highlighted expression is wrongly collected. The correct formulae should be $E_1=D_0(H_2)-D_0(H_2^+)+I(H)$.}
    \label{fig:p_error}
\end{figure*}

To help better understand the error category listed in Section~\ref{sec:error_analysis}, we provide one example for each category in Figure~\ref{fig:cal_error}, Figure~\ref{fig:p_error}.

